\documentclass[lettersize,journal]{IEEEtran}
\usepackage{amsmath,amsfonts}
\usepackage{algorithmic}
\usepackage{algorithm}
\usepackage{array}
\usepackage[caption=false,font=normalsize,labelfont=sf,textfont=sf]{subfig}
\usepackage{textcomp}
\usepackage{stfloats}
\usepackage{url}
\usepackage{verbatim}
\usepackage{graphicx}
\usepackage{cite}
\begin{document}


\title{Technical Report: Masked Skeleton Sequence Modeling for Learning Larval Zebrafish Behavior Latent Embeddings}


\author{Lanxin Xu, Shuo Wang
\thanks{Manuscript received xx xxxx 2024; revised xx xxxx 2024; accepted xx xxxx 2024. Date of publication xx xxxx 2024; date of current version xx xxxx 2024. (Corresponding author: Shuo Wang.)}
\thanks{Lanxin Xu is with the Department of Neuroscience, Physiology and Pharmacology, University College London, London WC1E 6BT, UK (e-mail: laney.xu.21@ucl.ac.uk).}
\thanks{Shuo Wang is with the Department of XXXX, Harbin Institute of Technology, Harbin, Heilongjiang 150001, China (e-mail: wangshuohit900309@163.com).}
\thanks{Lanxin Xu and Shuo Wang contributed equally to this work.}
}

\markboth{Journal of \LaTeX\ Class Files,~Vol.~14, No.~8, August~2021}%
{Shell \MakeLowercase{\textit{et al.}}: Masked Sequence Modeling for Learning Larval Zebrafish Behavior Latent Embeddings}

\IEEEpubid{0000--0000/00\$00.00~\copyright~2021 IEEE}

\maketitle

\begin{abstract}
In this report, we introduce a novel self-supervised learning method for extracting latent embeddings from behaviors of larval zebrafish. Drawing inspiration from Masked Modeling techniquesutilized in image processing with Masked Autoencoders (MAE) \cite{he2022masked} and in natural language processing with Generative Pre-trained Transformer (GPT) \cite{radford2018improving}, we treat behavior sequences as a blend of images and language. For the skeletal sequences of swimming zebrafish, we propose a pioneering Transformer-CNN architecture, the Sequence Spatial-Temporal Transformer (SSTFormer), designed to capture the inter-frame correlation of different joints. This correlation is particularly valuable, as it reflects the coordinated movement of various parts of the fish body across adjacent frames. To handle the high frame rate, we segment the skeleton sequence into distinct time slices, analogous to "words" in a sentence, and employ self-attention transformer layers to encode the consecutive frames within each slice, capturing the spatial correlation among different joints. Furthermore, we incorporate a CNN-based attention module to enhance the representations outputted by the transformer layers. Lastly, we introduce a temporal feature aggregation operation between time slices to improve the discrimination of similar behaviors. 

\end{abstract}


\section{Introduction}
\IEEEPARstart{B}{ehavior} analysis is a fundamental research methodology in neuroscience, particularly in the study of animals. Conventional behavior analysis methods heavily depend on feature engineering, leveraging computer vision algorithms to identify animal instances within experimental settings and subsequently track their movement trajectories and skeletons. Manually selected features, such as the maximum moving speed, can be calculated from animal skeleton data. This allows researchers to extract valuable insights into the intricate nuances of animal behavior, enhancing the classification of behaviors and leading to a deeper understanding of their neural foundations and overall behavioral patterns.

Deep learning-based approaches for behavior classification primarily rely on supervised learning, where large-scale, manually-annotated behavior datasets play a crucial role in achieving high training performance. However, models trained solely on supervised signals often lack generalization, necessitating the development of new models for novel animal species or specialized research purposes. Designing experiments, collecting data, and conducting manual or semi-automatic annotations entail substantial effort. Furthermore, the effectiveness of feature engineering heavily relies on researchers' expertise and experiences, potentially introducing a subjective criterion.

Traditional approaches of classifying zebrafish behaviour usually involve crafting high-dimensional features comprising kinetic parameters——such as velocity, acceleration, angular velocity, and distance traveled——along with the duration and frequency of behaviors and skeletal characteristics \cite{MARQUES2018181}\cite{Jouary052324}. However, behavioral analysis based solely on handcrafted features often overlooks the temporal dynamics and spatial correlations among different joints across consecutive frames.

In summary, while supervised learning with annotated datasets is integral to deep learning-based behavior classification, the limitations of handcrafted features underscore the need for methodologies that effectively incorporate temporal information and spatial relationships for more robust and generalized model performance.

We introduced a novel approach known as the Masked Skeletal Sequence Autoencoder (MSAE), designed to extract meaningful behavioral representations from unlabeled skeleton sequence data obtained from larval zebrafish swimming recordings. The MSAE architecture is founded upon the Sequence Spatial-Temporal Transformer (SSTFormer), which enables the capturing of correlations among various joints in both short-term consecutive frames and the overarching long-term sequence. In our commitment to transparency and collaboration, we plan to release the source code for MSAE on GitHub. Furthermore, comprehensive details and results, based on public datasets, will be elaborated in our official paper, providing a thorough understanding of the methodology and its outcomes.

\section{Method}
In this section, we will briefly introduce the MSAE architecture and the modules used in it. 

\begin{figure*}[!ht]
\centering
\includegraphics[width=0.9\textwidth]{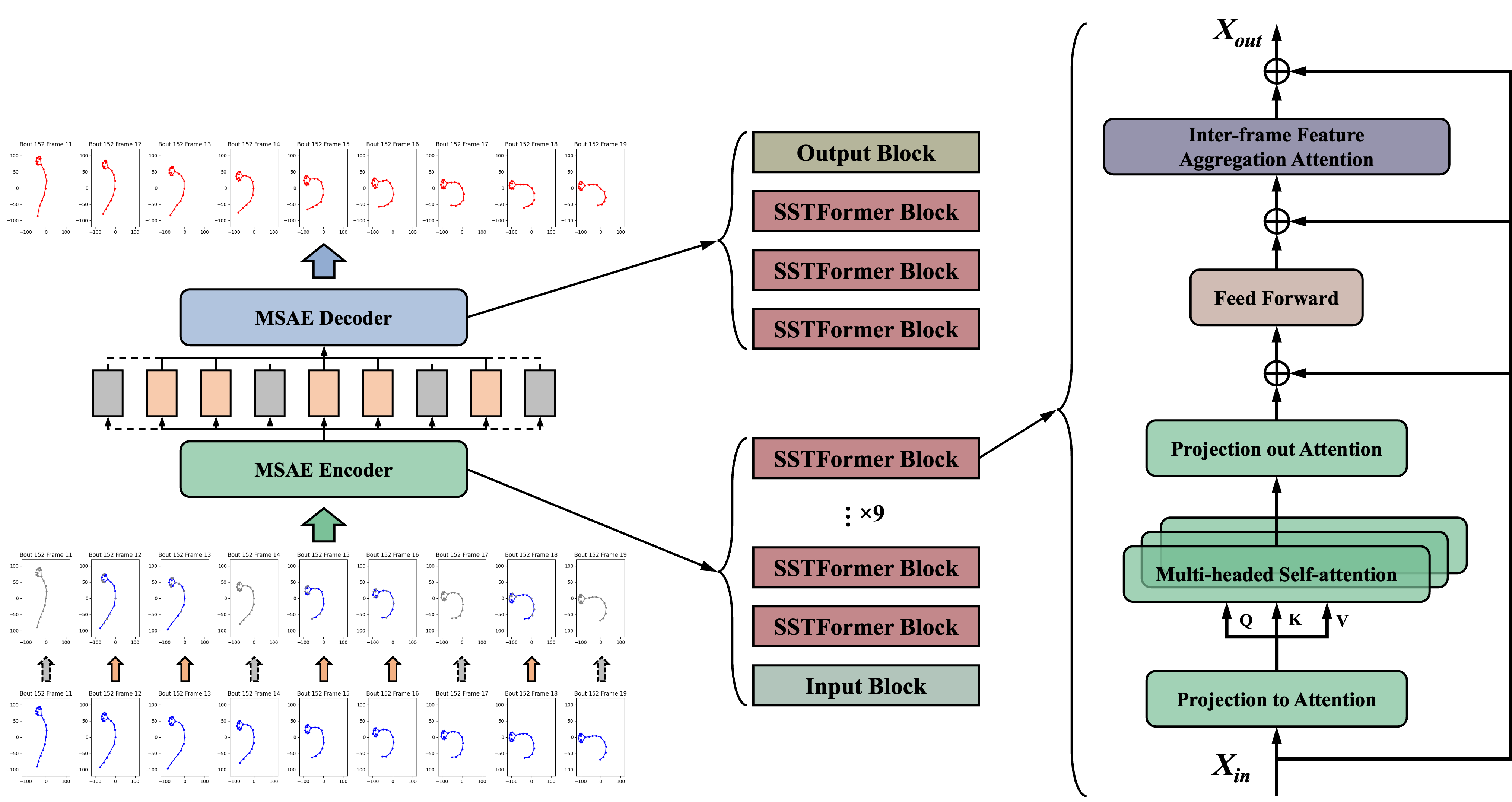}
\caption{The overview of the Masked Skeleton Sequence Autoencoder for the Larval Zebrafish swimming behavior.}
\label{msae}
\end{figure*}

Following the MAE design paradigm, the MSAE is crafted to embody a simple, efficient, and scalable manifestation of a masked skeleton sequence autoencoder designed for acquiring latent embeddings of larval zebrafish behavior in a self-supervised manner. The proposed MSAE employs random masking of joints and frames within the input skeleton sequence, thereafter reconstructing the absent joints and frames. As illustrated in Fig. \ref{msae}, notable is the discrepancy in size between the encoder and decoder, with the encoder being larger. Initially, the encoder processes the visible joints within the visible frames, subsequently allowing the decoder to reconstruct the input from the latent embeddings, inclusive of the masked tokens.

The MSAE could learn powerful models with strong generalization capabilities. Due to MSAE, we can train data-hungry models on large unlabeled skeleton sequence dataset. The pre-trained models theoretically can improve models' generalization performance and will be the better choices for transfer learning to improve downstream tasks. We think some other similar observations will be found in the further research and will be aligned with those witnessed in self-supervised pre-training in NLP \cite{radford2018improving,radford2019language,brown2020language,devlin2018bert} and CV \cite{he2022masked}.

\begin{figure}[!ht]
\centering
\includegraphics[width=3.2in]{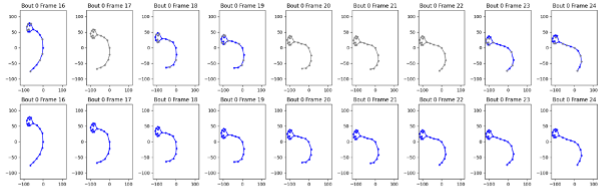}
\caption{The spatial-temporal encoding block.}
\label{sstm}
\end{figure}

\begin{figure}[!ht]
\centering
\includegraphics[width=2.0in]{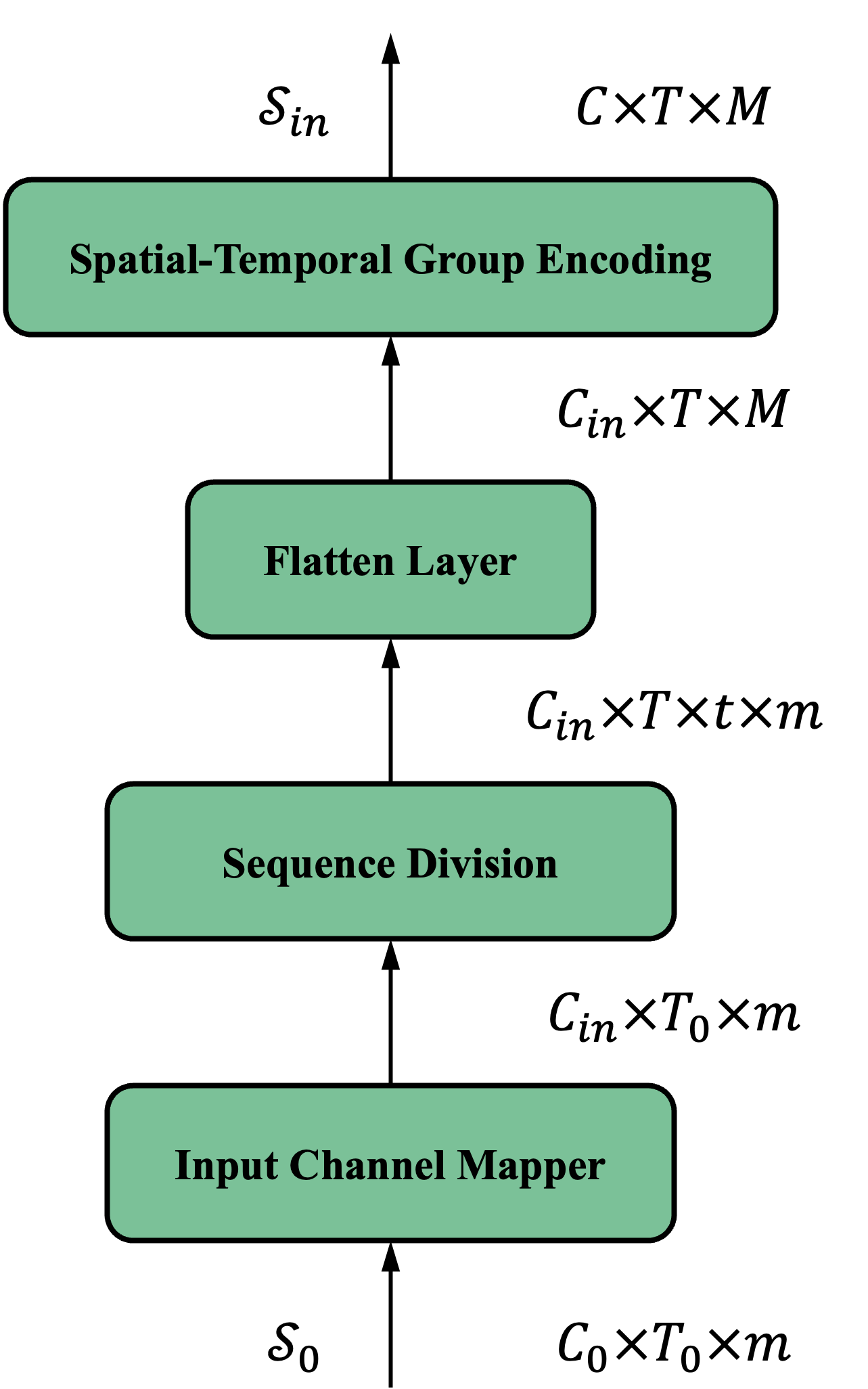}
\caption{The spatial-temporal encoding block.}
\label{stfe}
\end{figure}

\textbf{Masking}. The overall process of the Sequence Spatial-temporal Masking Strategy (SSTM) proposed in this study is illustrated in Fig. \ref{sstm}. The SSTM strategy consists of two stages: Temporal Masking and Spatial Masking. The first stage is Temporal Masking, which involves masking data at the frame level. Across the entire sequence of images, a certain percentage of frames are randomly removed according to a pre-defined masking ratio, and the indices of the masked frames are stored. The visible frames are then subjected to the second stage, Spatial Masking, where random masking is performed at the joint level based on pre-defined masking ratios. It is noteworthy that in this random spatial masking approach, the indices of masked joints are not fixed, meaning the same joint in different frames may be masked or remain visible. To model the relationships between different joints within consecutive frames, our work proposes a Spatio-Temporal Segment Encoding module (STSE) to encode masked joints. The structure of the encoding module is illustrated in Fig. \ref{stfe}.

\textbf{SSTFormer}. For skeletal sequence data, this study introduces a Sequence Spatial-Temporal Transformer (SSTFormer) structure, designed and constructed as the encoder and decoder networks for MSAE. SSTFormer is a skeleton data-driven Transformer that efficiently handles spatiotemporal information in skeletal sequence data. Specifically, SSTFormer segments skeletal sequence data into several non-overlapping segments and designs a Spatial-temporal Group Attention (STGA) mechanism for self-attention representation extraction, extracting multi-joint representations between adjacent frames. Furthermore, it integrates representations of inter-frame actions through the proposed Inter-frame Feature Aggregation (IFFA) module and further enhances the representation expressive power through convolution-based attention mechanism modules, thereby improving the learning capability for recognizing similar actions.

\begin{figure*}[!ht]
\centering
\includegraphics[width=0.9\textwidth]{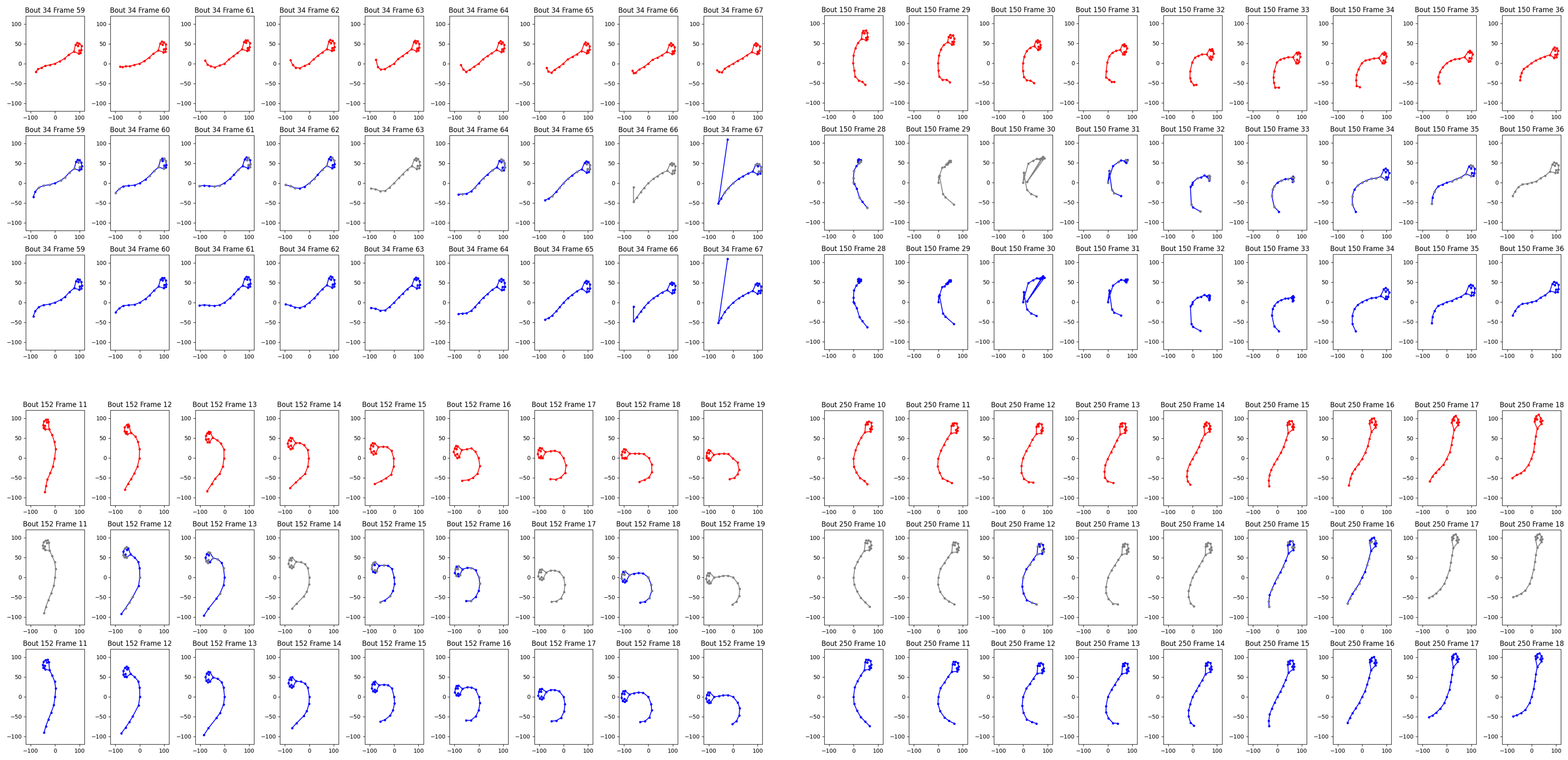}
\caption{The reconstruction results of MSAE on the validation set, where red denotes the reconstruction output by MSAE, blue represents the visible frames or joints according to MSAE, and grey indicates the invisible frames or joints.}
\label{results}
\end{figure*}

\textbf{MAE encoder}. Our encoder is composed of one STSE layer and 9 SSTFormer layers and applied only on visible, unmasked joints. 
The encoder embeds joints by a STSE layer with added positional embeddings, and then processes the resulting set via a series of STTFormer blocks. In MSAE, the encoder only operates on a small subset (e.g., $50\%$) of the full joints set. Masked joints and frames are removed; no mask tokens are used.

\textbf{MAE Decoder}. The MSAE decoder comprises 4 SSTFormer layers. It processes spatiotemporal embedding representations to reconstruct masked joints and skeletal frames, thus restoring the original skeletal sequence. To preserve positional information during skeleton sequence reconstruction, the decoder also incorporates the positional information of invisible joints and skeletal frames. The output of the decoder should closely resemble the original sequence. The MSAE decoder is exclusively used during pre-training for the skeleton sequence reconstruction task.

\textbf{Reconstruction Target}. MSAE reconstructs the input skeleton sequence by predicting the coordinates of joints for each masked joint. Each element in the decoder’s output is a vector representing a joint's coordinates in a specific frame. The last layer of the decoder is a linear projection with the number of output channels equal to the number of joints in a time slice. The decoder’s output is reshaped to form a reconstructed skeleton sequence. The loss function, mean squared error (MSE), is computed between the reconstructed and original joint coordinates.


\section{Experimental Results}
We used the skeleton sequence data provided by \cite{mullen2023poser}, all the information about this dataset can be found in their paper\cite{mullen2023poser}. The dataset provides 19 keypoints on the body of zebrafish larvae for each frame, with 4 points allocated for estimating eye keypoints, and the remaining 11 keypoints evenly distributed along the length from the nose tip to the tail end. Additionally, the dataset includes the segmentation of swimming bouts of zebrafish larvae, which can be directly utilized for the processing, training, and reconstruction of behavioral sequences. In total, the dataset comprises 34,015 swimming bouts.

The reconstruction results are shown in Fig. \ref{results}. Some interesting results will be described in future work, for example, MSAE can predict the correct coordinates of the joints that failed in the pose estimation by DLC\cite{Lauer2022MultianimalPE}.

\section{Conclusion and Discussion}
Thanks to the vigorous development of deep learning techniques, skeleton-based action recognition based on fully supervised learning has made significant progress. However, these methods require sufficient annotated data, which is not always readily available. In contrast, self-supervised skeleton-based action recognition has gained more attention. By leveraging a large amount of unlabeled data, more general representations can be learned to alleviate overfitting issues in skeleton-based action recognition tasks and reduce the demand for extensive annotated training data. Through self-supervised learning strategies in the MSAE paradigm pre-training, the encoder can generate temporally and spatially correlated universal skeleton representations.

In our ongoing research, after inputting complete skeleton sequences into the pre-trained encoder, we are attempting to integrate the generated universal skeleton representations with unsupervised clustering methods to explore behavior categories in unlabeled skeleton sequence data. Additionally, we are experimenting with using this pre-trained encoder for transfer learning to initialize supervised skeleton-based behavior prediction models. Following this, fine-tuning with a small amount of annotated data aims to achieve state-of-the-art performance.

\bibliographystyle{IEEEtran}
\bibliography{references}












\end{document}